\documentclass{elektr}
\usepackage{hyperref}
\hypersetup{
colorlinks=true,
urlcolor=blue,
citecolor=blue}
\usepackage[all]{xy,xypic}
\usepackage{amsfonts,amssymb,amsmath,amsgen,amsopn,amsbsy,theorem,graphicx,epsfig}
\usepackage{eufrak,amscd,bezier,latexsym,mathrsfs,eurosym,enumerate}
\usepackage[utf8]{inputenc}
\usepackage[english]{babel}
\usepackage{cleveref,multirow}
\usepackage[dvipsnames]{xcolor}
\usepackage[pagewise]{lineno}
\usepackage{natbib}
\usepackage{comment}

\usepackage{orcidlink}
\usepackage{colortbl}
\usepackage{adjustbox}
	
\definecolor{Gray}{gray}{0.9}

\usepackage{titlesec}

\titleformat{name=\section,numberless}
  {\normalfont\Large\bfseries}
  {}
  {0pt}
  {}

\yil{2021}
\vol{}
\fpage{}
\lpage{}
\doi{}

\title{Automated question generation and question answering from Turkish texts}

\author[Akyon et al]{
\textbf{Fatih Cagatay AKYON$^{1,2}$\thanks{fatih.akyon@metu.edu.tr}\orcidlink{0000-0001-7098-3944}~, Devrim CAVUSOGLU$^{1,3}$\orcidlink{0000-0002-5218-1283}~, Cemil CENGIZ$^{1}$\orcidlink{0000-0003-2681-5059}~}\\
\textbf{Sinan Onur ALTINUC$^{1,2}$\orcidlink{0000-0001-5119-160X}~, Alptekin TEMIZEL$^{2}$\orcidlink{0000-0001-6082-2573}}\\
$^{1}$OBSS AI, Ankara, Turkey\\
$^{2}$Graduate School of Informatics, METU, Ankara, Turkey\\
$^{3}$Computer Engineering, METU, Ankara, Turkey
\\ [1.8em]

\rec{.201}
\acc{.201}
\finv{..201}
}

\def\E{\ifmmode{\mathbb E}\else{$\mathbb E$}\fi} 
\def\N{\ifmmode{\mathbb N}\else{$\mathbb N$}\fi} 
\def\R{\ifmmode{\mathbb R}\else{$\mathbb R$}\fi} 
\def\Q{\ifmmode{\mathbb Q}\else{$\mathbb Q$}\fi} 
\def\C{\ifmmode{\mathbb C}\else{$\mathbb C$}\fi} 
\def\H{\ifmmode{\mathbb H}\else{$\mathbb H$}\fi} 
\def\Z{\ifmmode{\mathbb Z}\else{$\mathbb Z$}\fi} 
\def\P{\ifmmode{\mathbb P}\else{$\mathbb P$}\fi} 
\def\T{\ifmmode{\mathbb T}\else{$\mathbb T$}\fi} 
\def\SS{\ifmmode{\mathbb S}\else{$\mathbb S$}\fi} 
\def\DD{\ifmmode{\mathbb D}\else{$\mathbb D$}\fi} 

\newcommand{\bse}{\begin{subequations}}
\newcommand{\ese}{\end{subequations}}
\newcommand{\ben}{\begin{enumerate}}
\newcommand{\een}{\end{enumerate}}
\newcommand{\bens}{\begin{enumerate*}}
\newcommand{\eens}{\end{enumerate*}}
\newcommand{\be}{\begin{equation}}
\newcommand{\ee}{\end{equation}}
\newcommand{\bea}{\begin{eqnarray}}
\newcommand{\eea}{\end{eqnarray}}
\newcommand{\baa}{\begin{eqnarray*}}
\newcommand{\eaa}{\end{eqnarray*}}
\newcommand{\bc}{\begin{center}}
\newcommand{\ec}{\end{center}}

\theoremstyle{corollary}

\theoremstyle{lemma}

\theoremstyle{proposition}

\theoremstyle{axiom}

\theoremstyle{conjecture}

\theoremstyle{example}

\theoremstyle{definition}

\theoremstyle{remark}

\begin{document}

\maketitle

\begin{abstract}While exam-style questions are a fundamental educational tool serving a variety of purposes, manual construction of questions is a complex process that requires training, experience and resources. Automatic question generation (QG) techniques can be utilized to  satisfy the need for a continuous supply of new questions by streamlining their generation. However, compared to automatic question answering (QA), QG is a more challenging task. In this work, we fine-tune a multilingual T5 (mT5) transformer in a multi-task setting for QA, QG and answer extraction tasks using Turkish QA datasets. To the best of our knowledge, this is the first academic work that performs automated text-to-text question generation from Turkish texts. Experimental evaluations show that the proposed multi-task setting achieves state-of-the-art Turkish question answering and question generation performance on TQuADv1, TQuADv2 datasets and XQuAD Turkish split. The source code and the pre-trained models are available at \href{https://github.com/obss/turkish-question-generation}{https://github.com/obss/turkish-question-generation}.

\keywords{turkish, question answering, question generation, answer extraction, multi-task, transformer}
\end{abstract}

\section{Introduction}
\label{sec:intro}

Question Generation (QG) is the task of generating questions from a given context and, optionally, some answers. The research on QG has been developing exponentially with the task getting more popular in education \cite{kurdi2020systematic} \cite{lee2018automatic}, commercial applications such as chatbots and dialogue systems \cite{laban2021s} \cite{sreelakshmi2019question} and healthcare \cite{yue2020cliniqg4qa}.

Early works in QG were based mainly on human-designed sophisticated syntactic rules to transform a declarative sentence into the corresponding question. These tasks mainly relied on handcrafted feature extraction from documents. A method for generating multiple-choice tests from instructional documents (e.g., textbooks or encyclopedias) was proposed in \cite{mitkov2003computer}. In this work, domain-specific terms were extracted using the term frequency approach, and the sentences including the retrieved terms were transformed into questions using the parsed syntactic information of the sentence. In \cite{heilman2010good}, the input text is first simplified with a set of transformations to produce multiple declarative sentences. Then, a declarative sentence is transformed into a set of possible questions by syntactic and lexical transformations. However, being based on rule-based transformations, these methods are not applicable to other languages and question styles. A preliminary work provides an implementation plan for rule-based question generation from Turkish texts using syntactic (constituent or dependency) parsing and semantic role labeling systems \cite{soleymanzadeh2017domain}. In the QG part, manually generated templates and rules are used. However, the proposed method is not fully automated considering the manual selection of templates and its rule-based nature. Moreover, the paper does not provide sufficient technical details and no follow-up paper giving the details of the planned implementation is available.

Recently, many neural networks based techniques have been proposed for QG. An encoder-decoder architecture of an LSTM based seq2seq model is adopted in \cite{du2017learning}. Both the input sentence and the paragraph containing the sentence are encoded via separate bidirectional LSTMs \cite{hochreiter1997long} and then concatenated. This representation is then fed into the decoder, which is a left-to-right LSTM, to generate the question. The decoder learns to use the information in more relevant parts of the encoded input representation via an attention layer. Later models included target answer in the input to avoid questions that are too short and/or broadly targeted, such as "What is mentioned?". Some models have achieved that by either treating the answer’s position as an extra input feature \cite{zhou2017neural}, \cite{zhao2018paragraph} or by encoding the answer using a separate network \cite{duan2017question}, \cite{kim2019improving}. Moreover, position embeddings have been used to give more attention to the answer words closer to context words. Some utilized additional decoders to predict the question words (when, how, why, etc.) before generating the question \cite{sun-etal-2018-answer}. LSTM based Seq2seq models struggle to capture the paragraph-level context that is needed to generate high quality questions. Seq2seq model was extended with answer tagging, maxout pointer mechanism and a gated self-attention encoder \cite{zhao2018paragraph}. A multi-stage attention to link long document context with targeted answer is used in \cite{tuan2020capturing}.

%

Transformer based models have been dominating the NLP research in tandem with QG research lately. These models are capable of capturing longer and comprehensive contexts more effectively than their predecessors, mainly LSTM based seq2seq based models.
Named Entity Recognition (NER) is used in \cite{kriangchaivech2019question}, as a (preprocessing) task before the application of a transformer based model. In the work, they first extract a variety of named entities from the input text and then replace these entities with named entity tags for better generalization. 
Superior performances have been reported by applications of QG task that were proposed by those using a large transformer based language model (LM). A pre-trained BERTurk for QG was adopted in \cite{chan2019recurrent} and three models were proposed using BERTurk for QG, sequential question generation with BERTurk using the previous decoded results, and finally highlighting the answer in the context which yielded a performance improvement. A pre-trained GPT-2 for QG is used in \cite{lopez2020transformer} in a straightforward way by preparing inputs in a particular format. The model is evaluated in several ways such as context-copying, failures on constructing questions, robustness and answer-awareness.

An example question generation project using transformers in a specific framework is available\footnote{Neural question generation using transformers (2020). Website \href{https://github.com/patil-suraj/question_generation}{https://github.com/patil-suraj/question\_generation} [accessed 04 07 2021].} but it's sentence tokenization pipeline is specific to English language, presents the results on an English dataset, emphasizes limited input types (highlight and prepend) and does not have a peer reviewed publication.

Moreover, there is a publicly shared work based on fine-tuning mT5-small model \cite{xue2020mt5} on Turkish dataset for question generation task\footnote{Turkish Multitask MT5 (2021). Website \href{https://github.com/ozcangundes/multitask-question-generation}{https://github.com/ozcangundes/multitask-question-generation} [accessed 04 07 2021]} however it's sentence tokenization pipeline is not adapted to Turkish language, it is not clear whether the validation set is included in the training, does not present any evaluation results, emphasizes limited input types (only highlight) and does not have a peer reviewed publication. 

In this work, in order to fully automate the question generation process from Turkish texts using a single model, we propose a multi-task fine-tuning of mT5 model \cite{xue2020mt5}. To the best of our knowledge, this is the first comprehensive academic work that performs automated text-to-text question generation from Turkish texts. Main contributions can be summarized as, adaptation of a sentence tokenization pipeline for highlight input format, Turkish question generation and answering benchmarking of mT5 model on TQuADv1, TQuADv2 and XQuAD datasets in multi-task and single-task settings with different input formats (highlight/prepend/both).

\begin{figure}[!ht]
  \centering
  {\includegraphics[width=13.0cm]{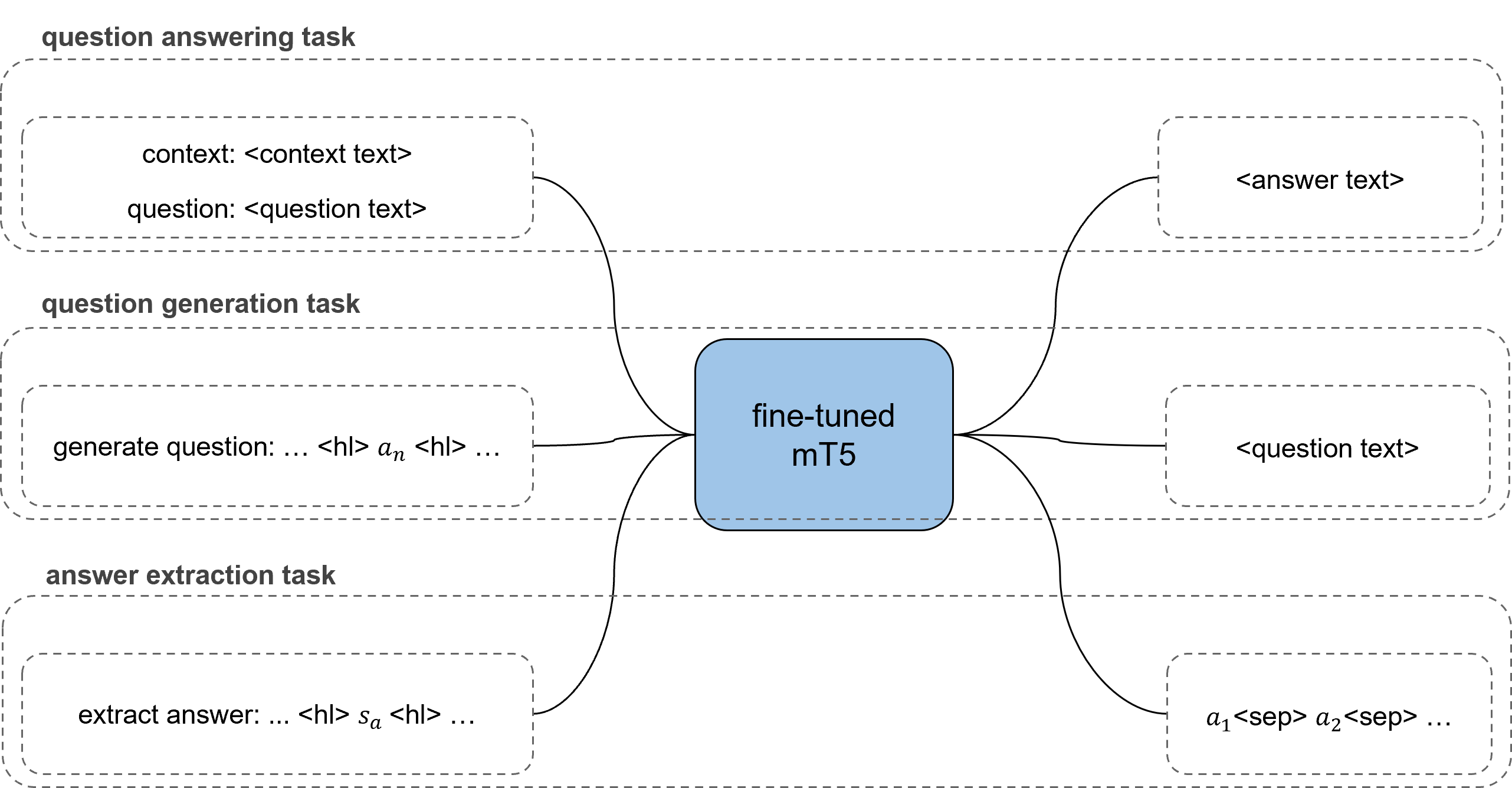}}
  \vspace{0.5cm}
    \vspace{0.5cm}
  {\includegraphics[width=12.0cm]{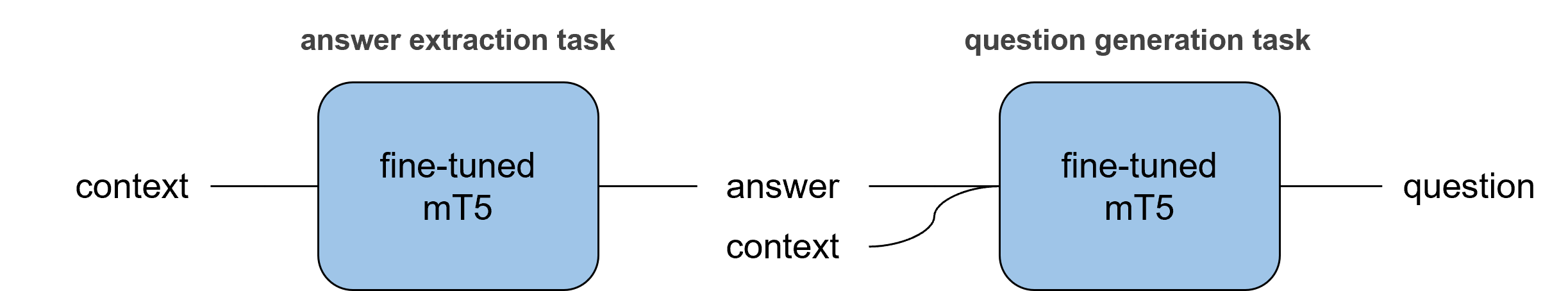}}
\caption{Multi-task fine-tuning of the multilingual pre-trained mT5 model (top). The same fine-tuned model is then used for both answer extraction and question generating task (bottom).}
\label{fig:mt-overall}
\end{figure}

The model we explored, mT5, is a variant of T5 \cite{raffel2020exploring}, which is a flexible Transformer model used in sequence-to-sequence NLP tasks. T5 is an encoder-decoder style language model whose architecture closely follows the original Transformer \cite{vaswani2017attention}. It is pre-trained on “span-corruption” objective, a special type of masked language modeling. In this scheme, consecutive input token spans are replaced with a mask token and the model is asked to reconstruct the original tokens in the spans as training objective. During fine-tuning, various distinct NLP tasks such as classification and generation formulated in common text-to-text format in multi-task learning setting. The main difference of mT5 is that it was trained on mC4 dataset, comprising natural text in 101 languages collected from the public Common Crawl web scrape. Being trained on multiple tasks and multiple languages, it can readily be fine-tuned on QA, QG and answer extraction tasks in Turkish language after converting the datasets to the common text-to-text format. As shown in Figure \ref{fig:mt-overall} (top), QA task uses context and question pair as input and answer as target, QG uses answer highlighted context as input and question as target, answer extraction uses sentence highlighted context as input and answer list with separator as target. This approach does not require an external answer extraction model or human effort to label the answers since the same model is used to extract the answers (corresponding to one of the potential questions) from the context as shown in Figure \ref{fig:mt-overall} (bottom).

\section{Proposed Approach}
\label{sec:proposed}

To convey a fully automated question generation pipeline, we assume that answer may not be given in the generation phase, and thus we also train the model to find answer $a$ (corresponding to one of the potential questions) which is a span in the given context $c$. The task is formulated as in Eq. \ref{Eq:Eq1} where $q$ denote the question targeting the answer $a$ and $c'$ is the context $c$ with highlighted tokens for sentences containing answers. 

\vspace*{-8px}
\begin{equation}
    P(q,a|c) = P(a|c')\cdot P(q|a,c)
    \label{Eq:Eq1}
\end{equation}

\hspace{-1cm}

 \textit{Answer extraction} task is formulated as $P(a|c')$ where \textit{context} and \textit{answer} pairs are used from \textit{\{context, question, answer\}} triplets from SQuAD style dataset. Context $c$ is first preprocessed to highlight the target answers, and the preprocessed context $c'$ is used as an input and answers are used as in training.

\textit{Question generation} task is formulated as $P(q|a,c)$ where \textit{\{answer, context\}} pairs are used as input and, in training, \textit{question} for the given answer is used as the target. If an answer is provided with the context, the answer extraction step is skipped, otherwise answer extraction is done before question generation. 

When providing the inputs to text-to-text transformer, different parts of the input \textit{c, a, q} are separated by a separator. In both single-task and multi-task QG setting, we apply three different input format styling: \emph{prepend}, \emph{highlight} and \emph{both}. In \emph{prepend} format, we prepend the base input text with a task specific prefix as in T5 \cite{raffel2020exploring}. For example, for QG task we prepend the base input format with $``generate \ question: "$ prefix. In \emph{highlight} format, we wrap the answer text in the context with a special \textit{\textless hl\textgreater \  (highlight)} token similar to \cite{xue2020mt5}. The \emph{both} input format contains both \emph{prepend} and \emph{highlight} input formats.





In the single-task setting, we modify each sample to train separate models for QA and QG tasks. In generation phase, QA task requires a context and a question and QG task requires a context and an answer as input. In the multi-task setting, we train the model to perform answer extraction, question generation and question answering tasks simultaneously. For answer extraction task, we put highlight tokens at the start and end of the sentence that contains the answer to be extracted. For the question generation, the answer of the question to be generated is highlighted  \cite{xue2020mt5}. Moreover, we prepend $``question"$, $``context"$, $``generate \ question"$, $``extract \ answer"$ tokens before each sample to help the model distinguish one task from other. In Figure \ref{fig:mt-overall}, the input and target formats of the model during fine-tuning is presented.

In single-task setting, for QG generation task, the answer always needs to be provided along with the context , whereas in multi-task setting the answer is not strictly required to be given.

 We adopt the answer-aware question generation methodology \cite{sun-etal-2018-answer}, where the model requires both the context and answer to generate questions. Use of the same model for automatic answer extraction in the multi-task setting eliminates the need for manual highlighting of the answer and enables end-to-end question generation from raw text. 
 
 It also has to be noted adapting the current schema to another language involves putting highlight tokens between sentences. However this might not be straightforward due to the language dependent nature of the sentence tokenization part, for which we needed to carefully design a proper sentence tokenization by manually handling edge cases mostly caused by abbreviations to mark the end of a sentence correctly in a Turkish text. There are wide options for sentence tokenization approaches for English text; however, there is no directly available sentence tokenization tool for Turkish text. We adapted an open-source tool TrTokenizer package\footnote{TrTokenizer (2020) Sentence and word tokenizers for the Turkish language \href{https://github.com/apdullahyayik/TrTokenizer}{https://github.com/apdullahyayik/TrTokenizer} [accessed 25.07.2020]} for a sentence tokenization step as the base tool and adapted it by enhancing the edge cases. These edge cases such as "Ar. Gör.", "(d. 998 - ö. 1068)", "Ömer b. Abdülazīz", etc. are then handled by regular expression based operations. The adapted Turkish sentence tokenization based answer highlighting, together with extended edge cases have been open-sourced in the project repo.



\section{Experimental Setup and Results}
\label{sec:exp}

We first fine-tune BERTurk \cite{stefan_schweter_2020_3770924} and mT5 models on TQuADv2 training split to have the base models.  Then F1, EM scores are calculated on TQuADv2 validation split and XQuAD Turkish split for experimental evaluation. All the experiments have been performed on Nvidia A100 GPU with 80 GB VRAM using Transformers Trainer \cite{wolf-etal-2020-transformers} on Pytorch \cite{paszke2019pytorch} backbone.

\subsection{Datasets}
\label{sec:datasets}

TQuAD \footnote{TQuad (2019) Turkish NLP Q\&A Dataset \href{https://github.com/TQuad/turkish-nlp-qa-dataset}{https://github.com/TQuad/turkish-nlp-qa-dataset} [accessed 04.07.2021]} (TQuADv1) is a Turkish QA dataset on Turkish \& Islamic Science History that was published within the scope of Teknofest 2018 Artificial Intelligence competition. TQuADv2 dataset \cite{tquadextended2021} extended the number of question-answer pairs along with the number of subjects by adding additional paragraphs and question-answer pairs to TQuADv1, i.e. TQuADv1 $\subset$ TQuADv2. Both of these datasets have the same structure with SQuAD \cite{rajpurkar2016squad}

XQuAD \cite{artetxe2019cross} is a multilingual QA dataset in Arabic, Chinese, German, Greek, Hindi, Russian, Spanish, Thai, Turkish and Vietnamese languages. It consists of samples  professionally translated from the SQuAD 1.1 validation set. The Turkish split of the XQuAD, namely XQuAD.tr, is used to evaluate the fine-tuned models, for brevity we denote it as XQuAD in the remainder of this paper.

The details of these datasets are provided in Table \ref{tab:datasets}, training sets are used for training the models and hyper-parameter tuning, validation set is used for performance evaluation only. Some examples are presented in Appendix \ref{sec:app}.

\begin{table}[!h]
    \caption{Details for datasets used in the proposed work.}
    \begin{center}
    \begin{tabular}{c|cc|cc}
    \hline
    \multicolumn{1}{c|}{Name} & \multicolumn{2}{c|}{Training set} & \multicolumn{2}{c}{Validation set} \\
    \newline & Paragraphs & QA-Pairs\newline & Paragraphs & QA-Pairs\newline \\
    \hline
    TQuADv1 & 2232 & 8308 & 275 & 892 \\
    TQuADv2 & 2400 & 14224 & 301 & 1330 \\
    XQuAD & - & - & 240 & 1190 \\
    \hline
    \end{tabular}
    \end{center}
    \label{tab:datasets}
\end{table}

\subsection{Hyper-parameter Tuning}
\label{sec:hyperparam}

\begin{table}[!h]
    \caption{QA scores of the best performing hyper-parameter combination for BERTurk, AdamW with initial learning rate of 1e-4 and 3 epocs.}
    \begin{center}
    \hspace*{0cm}
    \begin{tabular}{cccc}
    \hline
    TQuADv2-val\newline & TQuADv2-val\newline & XQuAD\newline & XQuAD\newline \\
    F1 & EM & F1 & EM \\
    \hline
    67.1 & 50.5 & 53.0 & 37.4 \\

    \hline
    \end{tabular}
    \end{center}
    \label{tab:bert-hyp}
\end{table}

\begin{table*}[!h]
    \caption{QA scores of the best performing hyper-parameter combination for mT5-small: AdamW with initial learning rate of 1e-3 and 15 epocs.}
    \begin{center}
    \begin{tabular}{cccc} 
    \hline

    TQuADv2-val\newline & TQuADv2-val\newline & XQuAD\newline & XQuAD\newline \\
    F1 & EM & F1 & EM \\
    \hline
    64.8 & 49.3 & 48.2 & 31.1  \\

    \hline
    \end{tabular}
    \end{center}
    \label{tab:mt5-hyp-qa}
\end{table*}

\begin{table*}[!h]
    \caption{QG scores of the best performing hyper-parameter combination for mT5-small: AdamW with initial learning rate of 1e-3 and 15 epocs.}
    \begin{center}
    \begin{tabular}{cccccc} 
    \hline
    TQuADv2-val\newline & TQuADv2-val\newline & TQuADv2-val\newline & XQuAD\newline & XQuAD\newline & XQuAD\newline\\
    BLEU-1 & BLEU-2 & ROUGE-L & BLEU-1 & BLEU-2 & ROUGE-L\\
    \hline

    39.3 & 32.8 & 45.6 & 21.3 & 14.0 & 28.5 \\

    \hline
    \end{tabular}
    \end{center}
    \label{tab:mt5-hyp-qg}
\end{table*}

We experimentally evaluated mT5 \cite{xue2020mt5} against BERTurk \cite{stefan_schweter_2020_3770924} and, to have a fair comparison, we performed hyper parameter tuning. For both models, we used grid-search to select the best optimizer type (Adafactor, AdamW), initial learning rate (\emph{1e-3}, \emph{1e-4}, \emph{1e-5}) and number of training epochs (1, 3, 5, 8, 10, 15, 20). BERTurk-base language model \cite{stefan_schweter_2020_3770924} has been fine-tuned for QA task on TQuADv2 training split, F1 and EM scores have been calculated on TQuADv2 validation split and XQuAD Turkish split. We selected the set of parameters which attain the overall best scores in all metrics: \emph{AdamW} optimizer with a learning rate of \emph{1e-4} and number of epochs 3 (shown in Table \ref{tab:bert-hyp}) for BERTurk. Similarly, mT5-small language model \cite{stefan_schweter_2020_3770924} has been fine-tuned in a multi-task setting on TQuADv2 training split. Then F1, EM scores for QA samples  and BLEU, ROUGE scores for QG samples have been calculated on TQuADv2 validation split and XQuAD Turkish split. QA and QG results of the best performing combination can be seen in Table \ref{tab:mt5-hyp-qa} and \ref{tab:mt5-hyp-qg}, respectively.
We selected the set of parameters which attain the overall best scores in all metrics: \emph{AdamW} optimizer with a learning rate of \emph{1e-3} and 15 epochs. For the remainder of the experiments, these fine-tuned BERTurk and mT5 models with the determined set of parameters have been used.

\subsection{Experimental Evaluation}

\begin{table}[!h]
    \caption{BERTurk-base and mT5-base QA evaluation results for TQuADv2 fine-tuning.}
    \begin{center}
    \begin{tabular}{c|cccc}
    \hline
    Setting & TQuADv2-val\newline & TQuADv2-val\newline & XQuAD\newline & XQuAD\newline \\
    & F1 & EM & F1 & EM \\
    \hline
    BERTurk & 67.1 & 50.5 & 53.0 & 37.4 \\
    Single-task mT5 & \textbf{71.6} & 55.1 & 60.7 & 40.2 \\
    Multi-task mT5 & 71.5 & \textbf{56.2} & \textbf{61.1} & \textbf{43.3} \\
    \hline
    \end{tabular}
    \end{center}
    \label{tab:qa-tquad2}
\end{table}

\begin{table}[!h]
    \caption{mT5-base QG evaluation results for single-task (ST) and multi-task (MT) for TQuADv2 fine-tuning.}
    \begin{center}
    \begin{tabular}{c|cccccc}
    \hline
    Setting & TQuADv2-val\newline  & TQuADv2-val\newline & XQuAD\newline & XQuAD\newline\\
        & BLEU-1 & ROUGE-L & BLEU-1 & ROUGE-L\\
    \hline
    MT-Both mT5 & \textbf{47.6} & \textbf{53.9} & \textbf{27.9} & \textbf{35.8} \\
    MT-Highlight mT5 & 45.9 & 52.5 & 26.2 & 34.8 \\
    MT-Prepend mT5 & 45.5 & 52.6 & 25.0 & 34.0 \\
    \hline
    ST-Both mT5 & 46.1 & 52.6 & 26.2 & 34.1 \\
    ST-Highlight mT5 & 45.2 & 52.4 & 25.8 & 33.5 \\
    ST-Prepend mT5 & 43.6 & 50.8 & 23.4 & 31.8 \\
    \hline
    \end{tabular}
    \label{tab:qg-tquad2}
    \end{center}
\end{table}

\begin{table}[!h]
    \caption{TQuADv1 and TQuADv2 fine-tuning QA evaluation results for multi-task mT5 variants and BERTurk. MT-Both means, mT5 model is fine-tuned with \textit{Both} input formats and in a multi-task setting.}
    \begin{center}
    \begin{tabular}{c|cccc}
    \multicolumn{5}{c}{\textbf{TQuADv1 fine-tuning results}} \\
    \hline
    Setting & TQuADv1-val\newline & TQuADv1-val\newline & XQuAD\newline & XQuAD\newline \\
    & F1 & EM & F1 & EM \\
    \hline
    BERTurk-base & 62.5 & 45.2 & 42.9 & 26.6 \\
    MT-Both mT5-small & 63.8 & 48.5 & 43.5 & 27.0 \\
    MT-Both mT5-base & 72.1 & 55.8 & 54.6 & 35.9 \\
    MT-Both mT5-large & 74.7 & 59.6 & 62.1 & 42.9 \\
    \hline
    \multicolumn{5}{c}{\textbf{TQuADv2 fine-tuning results}} \\
    \hline
    Setting & TQuADv2-val\newline & TQuADv2-val\newline & XQuAD\newline & XQuAD\newline \\
    & F1 & EM & F1 & EM \\
    \hline
    BERTurk-base & 67.1 & 50.5 & 53.0 & 37.4 \\
    MT-Both mT5-small & 65.0 & 49.3 & 48.8 & 32.9 \\
    MT-Both mT5-base & 71.5 & 56.2 & 61.1 & 43.3 \\
    MT-Both mT5-large & 73.3 & 58.4 & 65.0 & 46.7 \\
    \hline
    \end{tabular}
    \end{center}
    \label{tab:overall-qa}
\end{table}


\begin{table}[!h]
    \caption{TQuADv1 and TQuADv2 fine-tuning QG evaluation results for multi-task mT5 variants. MT-Both means, mT5 model is fine-tuned with 'Both' input format and in a multi-task setting.}
    \begin{center}
    \begin{tabular}{c|cccccc}
    \multicolumn{7}{c}{\textbf{TQuADv1 fine-tuning results}} \\
    \hline
    Setting  & TQuADv1-val\newline & TQuADv1-val\newline & TQuADv1\newline & XQuAD\newline & XQuAD\newline & XQuAD\newline\\
    & BLEU-1 & BLEU-2 & ROUGE-L & BLEU-1 & BLEU-2 & ROUGE-L\\
    \hline
    MT-Both mT5-small & 37.3 & 30.1 & 44.3 & 19.8 & 12.7 & 26.3 \\
    MT-Both mT5-base & 48.4 & 41.7 & 53.6 & 21.6 & 14.1 & 28.3 \\
    MT-Both mT5-large & 49.8 & 43.2 & 55.2 & 24.9 & 16.3 & 30.2  \\
    \hline
    \multicolumn{7}{c}{\textbf{TQuADv2 fine-tuning results}} \\
    \hline
    Setting & TQuADv2-val\newline & TQuADv2-val\newline & TQuADv2\newline & XQuAD\newline & XQuAD\newline & XQuAD\newline\\
    & BLEU-1 & BLEU-2 & ROUGE-L & BLEU-1 & BLEU-2 & ROUGE-L\\
    \hline
    MT-Both mT5-small & 39.6 & 32.9 & 46.5 & 21.1 & 13.8 & 28.4 \\
    MT-Both mT5-base & 47.6 & 41.2 & 53.9 & 27.9 & 20.9 & 35.8 \\
    MT-Both mT5-large & 49.1 & 42.7 & 54.3 & 29.3 & 21.9 & 37.5 \\
    \hline
    \end{tabular}
    \end{center}
    \label{tab:overall-qg}
\end{table}


\label{sec:results}

For the evaluation of QA task performance, widely accepted F1 and Exact Match (EM) scores \cite{rajpurkar2016squad} are calculated. Although there is no widely accepted automatic evaluation metric for measuring the QG performance\cite{amidei2018evaluation}, most of the previous works used the classical metrics like BLEU \cite{papineni2002bleu}, METEOR \cite{banerjee2005meteor} and ROUGE \cite{lin2004rouge}. METEOR applies stemming and synonym matching (in English). Hence, it has been excluded in our experiments as these processes are not applicable to Turkish. We reported BLEU-1, BLEU-2 and ROUGE-L metrics for evaluating the QG task performance. According to the TQuADv2 fine-tuning results in Table \ref{tab:qa-tquad2}, the proposed mT5 settings outperform the BERTurk in QA task and multi-task setting further increases the QA performance.

To evaluate the question generation performance of the proposed single and multi task settings, we fine-tuned mT5 model on TQuADv2 training split in both settings. Three different input formats explained in Section \ref{sec:proposed} are used and BLEU-1, ROUGE-L scores are calculated on TQuADv2 validation split and XQuAD Turkish split. According to TQuADv2 fine-tuning results in Table \ref{tab:qg-tquad2}, \emph{highlight} format increases the BLUE-1 scores by up to 1.6 points and ROUGE-L scores by up to 1.7 points compared to \emph{prepend} format in single-task setting. Moreover, \emph{highlight} format increases the BLUE-1 scores by up to 1.2 and ROUGE-L scores by up to 0.8 points compared to \emph{prepend} format in multi-task setting. Moreover, combining both techniques increases BLEU-1 scores by up to 2.9 points and ROUGE-L scores by up to 3.8 points compared to \emph{prepend} format.

Additional experiments have been conducted to evaluate the overall performance of the larger mT5 variants, mT5-base and mT5-large in comparison to BERTurk. QA and QG evaluation results for fine-tuned TQuADv1 and TQuADv2 are provided in Tables \ref{tab:overall-qa} and  \ref{tab:overall-qg} respectively. According to the QA results in Tables \ref{tab:overall-qa}, all mT5 variants outperform BERTurk  for smaller dataset sizes, BERTurk may outperform mT5-small for larger dataset sizes. This indicates that mT5 models are always preferable when the data is scarce whereas regular single-task training may also be used in place of the mT5-small variant when sufficient data is available. A comparative performance evaluation of mT5 variants shows that increasing the model size improves the performance significantly for both datasets, especially when switching from mT5-small to mT5-base. While using an even bigger model, mT5-large, improves the performance, it has a relatively more modest effect. Nevertheless, this trend of obtaining better scores by increasing the model capacity is consistent with the previous works on other transformer based models. Comparison of the results for different versions of the TQuAD datasets in Table \ref{tab:overall-qa} and \ref{tab:overall-qg} show that, although the TQuADv1 validation scores are higher than TQuADv2 validation scores, the models trained on the TQuADv2 train set are able to generalize better as indicated by the XQuAD Turkish split results. This can be attributed to the larger size and better quality of the TQuADv2 dataset.

For qualitative evaluation, some model outputs from different paragraphs are provided in Appendix \ref{sec:model-outputs} on Table \ref{tab:consistent-generation-samples} and \ref{tab:inconsistent-generation-samples} for consistent and inconsistent (that lacks coherence, not addressing the input answer, etc.) generations, respectively.

\section{Conclusions}
\label{sec:conc}
By combining the proposed answer extraction and answer-aware QG modules, it is possible to fully automate the QG task without any manual answer extraction labor. Automated evaluation metrics on TQuAD validation set show that the model is capable of generating meaningful question-answer pairs from the context after fine-tuning. Moreover, results show that the proposed multi-task approach has better performance on QA, answer extraction and QG compared to single-task setting. By combining the \emph{prepend} and \emph{highlight} input formats, QG performance of an mT5 model can be boosted up to 10\%.

In the future, further experiments with the multi-task model on the other QG tasks such as multiple choice, true/false, yes/no will be examined and effect of multilingual knowledge in mT5 will be analysed. In addition, human evaluations could be done to provide further insight about the performances of the methods.

\bibliography{refs}

\section*{Appendix}
\appendix

\section{Samples from Datasets}
\label{sec:app}

Here we provide some samples from datasets for visual inspection. For TQuADv1 (Table \ref{tab:tquadv1-samples}) and TQuADv2 (Table \ref{tab:tquadv2-samples}), samples are shown from both training and validation sets, and for XQuAD (Table \ref{tab:xquad-samples}), samples are shown with tag of validation set only as XQuAD is considered a validation set by itself as a whole. The answers are highlighted within the context with green background for ease of reading.

\begin{table}[!h]
    \caption{Some samples drawn from TQuADv1.}
    \begin{adjustbox}{width={\textwidth},totalheight={\textheight},keepaspectratio}%
    \begin{tabular}{c|c|p{15cm}}
    \hline
    Split & Sample & TQuADv1  \\
    \hline
    \multirow{24}{*}{Train} & \textbf{context} & Constantinus’un yaptığı etkiye gelince, Schipperges şu görüştedir: Avrupa tıbbına bir stratejik etkide bulunamamıştır. Constantinus’un külliyatı Solerno için çok önemli olsa da, Avrupa’daki diğer okullara sadece hazırlayıcı etkide bulunmuştur. Bu yargılamasında Schipperges, tıp kitapları resepsiyonunun ilk dalgasını, İber Yarımadası üzerinden gerçekleşen ikinci dalga ile karşılaştırdığı açısından haklı olabilir, şu kadar var ki, hazırlayıcı etkinin önemi küçümsenemez. Üstelik yalnızca bir tanesi müstesna olmak üzere, onun yaklaşık yirmi eserin çevirisi daha iyi çevirilerle yerlerini kaybetmiş değil, bilakis onlar yüzlerce yıl Constantinus’un kendi eserleriymiş gibi elden ele dolaşmıştır. Constantinus’un Latince’ye aktardığı eserlerle ilişkisi sözkonusu olduğunda Schipperges onu aşırmacı olarak nitelemekten kaçınmaktadır. Ona göre, Constantinus’un çalışmaları \colorbox{green}{alışıldık terim “resepsiyon” ile} nitelendirilemez, onun yaptığı daha çok başlangıçtan beri, belirli bir organik amaç için yabancı bilgi malzemesinin bilinçli bir koadunasyonu (bir öğretinin geniş bir kitle için yorumlanması/ şerhi) ve adaptasyon formunda işlenmesi olarak nitelendirilebilir. Bu faaliyet için asimilasyon/ özümseme daha doğru bir ifadedir68. Ama ben şahsen, Schipperges’in Constantinus’un Latince’ye aktardığı eserlerle olan ilişkisinin tarzına ve şekline yönelik yaptığı bu nitelemelerde isabetli davrandığına inanmıyorum. Bence Constantinus’un çevirilerinde söz konusu olan husus, kendine has bir resepsiyon tarzıdır. Constantinus’un, çevirdiği eserlerin gerçek yazarlarının isimlerini saklamaya asla hakkı yoktu. Bu durum karşısında onun neden böyle davrandığı sorusunu cevaplandırmak gerekir. 1930 yılında buna ilişkin olarak Hermann Lehmann şöyle demektedir: Constantinus’un bu davranışıyla Salerno’daki yüksek okulun gözündeki üstünlüğünü yüceltmek istemiş olacağından başka bir şey düşünemiyorum. Ben daha farklı bir açıklamaya varıyorum. Buna göre, Constantinus’un kendisine esas aldığı eserlerle olan bu aşırmacı ilişkisi birden çok faktörle izah edilebilir:\\
    & \textbf{question} & \textbf{Schippinges’e göre Constantinus’un çalışmaları hangi terim ile nitelendirilemez?}\\
    & \textbf{answer} & alışıldık terim “resepsiyon” ile\\
    \hline
    \rowcolor{Gray}
    \multirow{14}{*}{Train} & \textbf{context} & İslam dünyasında bilimin 16. yüzyılda hala yüksek seviyede bulunduğunu gösteren çok ilginç bir örneği deskriptif coğrafya ekolünden verebiliriz. Bize bu örneği, Avrupa’da \colorbox{green}{Afrikalı Leo} (Leo Africanus) olarak tanınan el-hasan b. Muhammed el-Vezzan (doğumu yaklaşık 888/1483)’dır. Fas (Fez) şehrinde büyümüş ve eğitimini almış olan Granada doğumlu bu bilgin, diplomatik hizmetler yoluyla, özellikle kuzey Afrika’da olmak üzere birçok İslam ülkesini tanıyıp bir yazar olarak coğrafya ve kartografya ile ilgileniyordu. İstanbul’dan dönüş yolculuğunda Sicilyalı korsanların eline esir düşmüş, ilk olarak Napoli’ye daha sonra Roma’ya satılıp Papa X. Leo tarafından 6.1.1520 yılında bizzat Papa’nın adıyla Giovanni Leo olarak vaftiz edilmişti. İtalya’daki ikameti sırasında İtalyanca öğrendi ve Arapça öğretti. Yazar olarak faaliyetlerini Roma ve Bologna’da devam ettirdi. Afrika coğrafyası dışında kuzey Afrikalı 30 bilginin biyografilerini içeren diğer bir eser derledi. Afrika kitabını esaretinin 6. yılı olan 1526’da İtalyan dilinde tamamladı. 935/1529 yılında Tunus’a döndü ve orada Müslüman olarak öldü.\\
    \rowcolor{Gray}
    & \textbf{question} & \textbf{el-Hasan b. Muhammed el-Vezzan isimli bilgin avrupa’da nasıl tanınmaktadır ?}\\
    \rowcolor{Gray}
    & \textbf{answer} & Afrikalı Leo\\
    \hline
    \multirow{12}{*}{Val} & \textbf{context} & Pardus için mevcut işletim sistemleri, başta Linux olmak üzere incelendi, açık kaynak yazılım metodolojisi (yöntem bilimi) ve felsefesi ayrıntılı olarak çalışıldı. Bu incelemeler sonrasında, 2003 yılı sonbaharında, Linux temelli, açık kaynaklı, olabildiğince GPL lisanslama yöntemini kullanan bir işletim sistemi dağıtımı oluşturulmasına karar verildi. Pardus Projesi'nin hayata geçmesi, 2004 yılı başında teknik ekibin çekirdeğinin oluşturulmasıyla başladı. Bu aşamada Türkiye'nin Linux geçmişi, mevcut ve planlanan dağıtımlar, açık kaynak ve Linux camiası ve girişimleri de göz önüne alınarak, var olan bilgi birikimi ve deneyimden en üst düzeyde yararlanmanın yolları arandı. Sonuçta ulusal işletim sistemi geliştirilmesinde görev alması en uygun kişiler Türkiye'nin dört bir yanından seçilerek \colorbox{green}{TÜBİTAK/UEKAE} bünyesinde bir araya geldiler.\\
    & \textbf{question} & \textbf{Pardus’un 2004 yılı başlarında hayata geçirilen projelerinde çalışacak en uygun kişiler hangi kuruluşun çatısı altında toplanmışlardır?}\\
    & \textbf{answer} & TÜBİTAK/UEKAE\\
    \hline
    \end{tabular}%
    \end{adjustbox}
    \label{tab:tquadv1-samples}
\end{table}

\begin{table}[!h]
    \caption{Some samples drawn from TQuADv2.}
    \begin{adjustbox}{width={\textwidth},totalheight={\textheight},keepaspectratio}%
    \begin{tabular}{c|c|p{15cm}}
    \hline
    Split & Sample & TQuADv2  \\
    \hline
    \multirow{12}{*}{Train} & \textbf{context} & Hünkâr İskelesi Antlaşması ile iki devlet arasında oluşan ittifakın hâtırası olmak üzere \colorbox{green}{Rus askerlerinin ordugâh kurdukları yere} bir anıt dikilmiştir (zamanında yapılmış gravürü için bk. Kutluoğlu, rs. nr. 7, 8). Anıtın iki cephesinde Türkçe ve Rusça olmak üzere iki devlet arasındaki dostluğu tebcil eden mısralar, İstanbul halkının bu gelişme karşısında duyduğu infiali aksettirmemektedir. Muhafazakâr çevrelerin Rusya’ya karşı duymakta olduğu nefret ve özellikle II. Mahmud aleyhine sarfettikleri sözler, Avusturya elçisi Baron von Stürmer’in raporlarında görülen ve padişahı “sarhoş, çılgın, gâvur” olarak niteleyen kayıtlarla da sabittir. Rus ittifakına karşı duyulan bu infialin arkasında, Mehmed Ali Paşa’nın bütün Avrupa’da pek etkili olan ve büyük para gücü ile devreye sokup başarı ile yürüttüğü propaganda faaliyetlerinin olduğu açıktır.\\
    & \textbf{question} & \textbf{Hünkâr İskelesi Antlaşması ile iki devlet arasında oluşan ittifakın hâtırası olmak üzere nereye bir anıt dikilmiştir ?}\\
    & \textbf{answer} & Rus askerlerinin ordugâh kurdukları yere\\
    \hline
    \rowcolor{Gray}
    \multirow{9}{*}{Train} & \textbf{context} & Ferhat Paşa Antlaşması, III. Murat döneminde, 21 Mart 1590 tarihinde, Osmanlı Devleti'yle Safevi Devleti arasında imzalanmış bir barış antlaşmasıdır. Ferhat Paşa Antlaşması, 1578-1590 tarihlerindeki Osmanlı-Safevî Savaşı'nı sona erdirmiştir. Ferhat Paşa Antlaşması duraklama döneminin ilk antlaşmasıdır. Ferhat Paşa Antlaşması \colorbox{green}{İstanbul'da} imzalanmıştır. Ferhat Paşa Antlaşması sonucunda Tebriz, Karabağ, Gürcistan, Dağıstan ve Şirvan Osmanlılara bırakıldı. Ferhat Paşa Antlaşması ile Osmanlılar doğudaki en geniş sınırlarına ulaşmışlardır. Ferhat Paşa Antlaşması, III. Mehmet döneminde, Safeviler tarafından ihlal edilmiştir.\\
    \rowcolor{Gray}
    & \textbf{question} & \textbf{Ferhat Paşa Antlaşması nerede imzalanmıştır?}\\
    \rowcolor{Gray}
    & \textbf{answer} & İstanbul'da\\
    \hline
    \multirow{28}{*}{Val} & \textbf{context} & Mehmed Ali Paşa’nın karşı propagandasından ve genel bir infialden çekinen Bâbıâli’nin filonun Boğaz dışında Süzeboli açıklarında demir atması teklifleri ise etkisiz kaldı. Rus filosunun Boğaz’a gelmesi karşısında ne Fransa’nın Mehmed Ali Paşa yanlısı politikasında, ne de İngiltere’nin ısrarla sürdürmekte olduğu tesbit edilen ilgisiz tutumunda bütün diplomatik teşebbüslere rağmen herhangi bir değişiklik görmeyen Bâbıâli, başşehrin müdafaası için Rusya kara kuvvetlerinin de getirilmesine karar verdi (1833 Mart sonları). 5000 kişilik bir Rus kuvveti 5 Nisan’da İstanbul’a gelerek Beykoz’da karaya çıkıp Hünkâr İskelesi’nde karargâh kurdu. O sıralarda Mısır kuvvetleriyle anlaşmak üzere Kütahya’da görüşmeler sürdürülmekteydi. Nihayet Adana’nın da “muhassıllık” olarak İbrâhim Paşa’ya bırakılması, Mısır ve bütün Suriye vilâyetlerinin, diğer bir ifadeyle I. Selim’in 1516 Mercidâbık ve 1517 Ridâniye zaferleriyle ele geçen bütün yerlerin, Mehmed Ali Paşa’ya terkiyle varılan uzlaşma neticesinde Mısır meselesinin birinci safhası sona ermiş ve bu tevcîhatı ihtiva eden ferman 6 Mayıs’ta Mehmed Ali Paşa’ya gönderilerek iki taraf arasındaki ihtilâfa şimdilik son verilmişti. Mısır kuvvetlerinin Toroslar’ın öte taraflarına çekilmesi işi başlamış olmakla beraber Rus kara ve deniz kuvvetlerinde herhangi bir toparlanma gözlenmediği gibi, 5 Mayıs 1833’te çarın büyük yetkilerle yolladığı olağan üstü elçisi Aleksey Orlof’un (Alexej Orlow) İstanbul’a gelişi yapılan bu yardımın faturasının ağır olacağına işaret etmekteydi. Nitekim Orlof, iki devlet arasında 3 Ocak 1799’da yapılan antlaşmaya benzer bir savunma ittifakı teklif etmekte gecikmemiştir. Öte yandan Rusya’nın kara ve deniz kuvvetlerini geri çekmemesi İngiltere ve Fransa’nın donanma nümayişlerine yol açmaktaydı. Ortak harekât, Boğazlar’dan geçerek Rus kuvvetlerini geri çekilmeye zorlama teşebbüslerine kadar varan temayüller gösterdiyse de Rus ittifak teklifinin Bâbıâli tarafından reddine yeterli olmadı. İlk esasları Ahmed Fevzi Paşa ve Orlof arasında tesbit edilen görüşmelere daha sonra Serasker Hüsrev Paşa ve Reîsülküttâb Âkif Efendi ile Rus elçisi Butenef de katıldı. İlk toplantı 26 Haziran’da \colorbox{green}{Hüsrev Paşa’nın Emirgân’daki yalısında} yapıldı ve 8 Temmuz’daki ikinci toplantıda antlaşma burada imzalandı. Antlaşmanın yapılmasından iki gün sonra Rus filosu ve sayıları 13 veya 15.000’e varan kara kuvvetleri Karadeniz’e çıkmak üzere Boğaz’dan hareket etti.\\
    & \textbf{question} & \textbf{8 Temmuz’daki ikinci toplantıda antlaşma nerede imzalandı ?}\\
    & \textbf{answer} & Hüsrev Paşa’nın Emirgân’daki yalısında\\
    \hline
    \end{tabular}%
    \end{adjustbox}
    \label{tab:tquadv2-samples}
\end{table}

\begin{table}[!h]
    \caption{Some samples drawn from XQuAD.}
    \begin{adjustbox}{width={\textwidth},totalheight={\textheight},keepaspectratio}%
    \begin{tabular}{c|c|p{15cm}}
    \hline
    Split & Sample & XQuAD  \\
    \hline
    \multirow{6}{*}{Val} & \textbf{context} & Peyton Manning,iki farklı takımı birden fazla Super Bowls'a gitmesinde önderlik eden ilk kilit oyuncu oldu. Ayrıca, 39 yaşında bir Super Bowl'da oynayan en eski oyun kurucu oldu. Geçmiş rekor, Broncos'u 38 yaşında Super Bowl XXXIII'te galibiyete götüren ve şu anda Denver'ın Futbol Operasyonları Başkan Yardımcısı ve Genel Müdür'ü olan \colorbox{green}{John Elway}'in ellerindeydi.\\
    & \textbf{question} & \textbf{Bir Super Bowl'da oynayan en yaşlı kilit oyuncu olma rekorunu daha önce kim elinde bulunduruyordu?}\\
    & \textbf{answer} & John Elway\\
    \hline
    \rowcolor{Gray}
    \multirow{13}{*}{Val} & \textbf{context} & Polonya'daki bölgesel ayrılma temel birimi bir komündür  (gmina). Bir şehir aynı zamanda bir komündür - ancak kent tüzüğü ile. Hem şehirler hem de komünler bir belediye başkanı tarafından yönetilmektedir - ancak komünlerde belediye başkanı vogt'tur (Lehçe'de wójt), ancak şehirlerde - burmistrz'dir. Bazı daha büyük şehirler, bölgedeki bölünmenin ikinci seviyesindeki birimler tarafından sahip olunan görev ve ayrıcalıklar gibi yetkileri elde ederler - \colorbox{green}{ilçeler veya powiat}'lar. Bu yetkilere örnek olarak bir otomobil tescili verilebilir: bir gmina otomobilleri kaydedemez, bu bir powiat'ın görevidir (yani bir kayıt numarası, bir arabanın gmina'ya değil, hangi powiat'a kaydedilmiş olduğuna bağlıdır). Bu durumda biz şehir, ilçe veya powiat grodzki hakkında konuşuyoruz. Bu şehirler örneğin Lublin, Krakov, Gdansk, Poznan'dır. Varşova'da, ilçelerinde ek olarak bazı powiat'ların yetkileri bulunmaktadır - daha önce belirtilen araba kaydı gibi. Örneğin, Wola ilçesinin kendi kanıtı vardır ve Ursynów bölgesi- kendi (ve Wola'nın otomobillerinin Ursynów’unkinden başka bir kayıt numarası) vardır. Ancak, örneğin Kraków'daki ilçelerde powiat hakları yoktur, bu nedenle Kraków'daki kayıt numaraları tüm ilçelerde aynı tiptedir.\\
    \rowcolor{Gray}
    & \textbf{question} & \textbf{Polonya'daki bölgesel ayrımlardaki ikinci seviye nedir?}\\
    \rowcolor{Gray}
    & \textbf{answer} & ilçeler veya powiat\\
    \hline
    \end{tabular}%
    \end{adjustbox}
    \label{tab:xquad-samples}
\end{table}

\section{Model Outputs}
\label{sec:model-outputs}

Here we provide some sample question generation results from TQuADv2 dataset for visual inspection. For sample consistent (Table \ref{tab:consistent-generation-samples}) and inconsistent (Table \ref{tab:inconsistent-generation-samples}) question generations, results are shown. The answers are highlighted within the context with green background for ease of reading.

\begin{table}[!h]
    \caption{Some consistent question generation results from TQuADv2.}
    \begin{adjustbox}{width={\textwidth},totalheight={\textheight},keepaspectratio}%
    \begin{tabular}{c|c|p{15cm}}
    \hline
    Split & Type & Text  \\
    \hline
    \multirow{7}{*}{Val} & \textbf{context} & Turkcell , Türkiye merkezli teknolojik iletişim operatör şirketidir. GSM, 2G , 3G , 4G ve 4.5G operatörüdür. \colorbox{green}{GSM 900, UMTS2100, LTE800, LTE900, LTE1800, LTE2100, LTE2600} teknolojilerini kullanarak hizmet vermektedir. Kurulduğu günden bu yana, lisans bedeli de dahil olmak üzere, yurt içerisinde, Turkcell ileriye dönük hedeflerinde 18 milyar Lira yatırım yapmayı hedeflemiştir ve yüzbinlerce Türk vatandaşına iş istihdamı sağlamıştır.\\
    & \textbf{generated question} & \textbf{Turkcell hangi frekanslar üzerinden hizmet vermektedir?}\\
    & \textbf{gold answer} & GSM 900, UMTS2100, LTE800, LTE900, LTE1800, LTE2100, LTE2600\\
    \hline
    \rowcolor{Gray}
    \multirow{12}{*}{Val} & \textbf{context} & Turkcell \colorbox{green}{11 Temmuz 2000} yılından itibaren İstanbul borsasında hisselerini satışa çıkardığından beri New York Borsası'nda listelenen ilk Türk şirketi olmuştur. Turkcell'in hissedar yapısı aşağıdaki gibidir: \%51 Turkcell Holding A.Ş'ye ait, \%0.05 Çukurova Holding A.Ş'ye, \%13,07 Sonera Holding B.V'ye, \%1,18 MV Holding A.Ş'ye aittir ve serbest payda ise \%34,7 dir. 2011'in Aralık ayında Sonera Holding ve Çukurova Grubu doğrudan ve dolaylı olarak sırasıyla Turkcell'in hisse senedinin yaklaşık \%37.1 ve \%13.8'ini sahiplendiler. Çukurova Grubu 2005 yılının mart ayında TeliaSonera'ya hissesinin büyük bir kısmını sattı. O zamandan beri bu iki firma anlaşıp anlaşamama konusunda tartışıyorlar. 2009 yılının Ağustos ayında Uluslararası Ticaret Odası Çukurova'nın Turkcell Holding içindeki kalan tüm hisselerini TeliaSonera‘ya dağıtmasının gerektiği durumunda bir hüküm yayınladı.\\
    \rowcolor{Gray}
    & \textbf{generated question} & \textbf{Turkcell hisselerini İstanbul borsasında hangi tarih itibariyle satışa çıkarmıştır?}\\
    \rowcolor{Gray}
    & \textbf{gold answer} & 11 Temmuz 2000\\
    \hline
    \multirow{10}{*}{Val} & \textbf{context} & Bu kitaplardan Tarih-i Çelebizade 'nin pek çok nüshası Tarih-i Raşid 'in üçüncü ve son cildinin arkasına eklenip ciltlenerek beraber satıldığı için, bazı kaynaklar hataya düşerek Müteferrika'nın bastığı kitap sayısını 16 olarak göstermektedirler. Basılan kitapları çoğunlukla Kethüda, Mektupçu, Çavuşbaşı gibi Osmanlı Bürokrasisi ve Şeyhülislam, Kadı v.b. satın almıştır. Özellikle ilmiye sınıfından kişilerin kitapları alması ulemanın matbaaya karşı olmadığını göstermektedir. Buna karşın \colorbox{green}{satış adetleri çok düşük kaldığından} İbrahim Müteferrika , latince olarak bastırdığı kataloglarla Avrupa'nın değişik yerlerinde kitaplarını satmaya çalıştı. Örneğin Grammaire Turque den 200 adedini Cizvit Mektebine, peşin fiyatı 3 kuruş iken toptan 2,5 kuruşa satmıştır.\\
    & \textbf{generated question} & \textbf{İbrahim Müferrika neden kitaplarını satmaya çalıştı?}\\
    & \textbf{gold answer} & satış adetleri çok düşük kaldığından\\
    \hline
    \end{tabular}%
    \end{adjustbox}
    \label{tab:consistent-generation-samples}
\end{table}

\begin{table}[!h]
    \caption{Some inconsistent question generation results from TQuADv2.}
    \begin{adjustbox}{width={\textwidth},totalheight={\textheight},keepaspectratio}%
    \begin{tabular}{c|c|p{15cm}}
    \hline
    Split & Type & Text  \\
    \hline
    \multirow{10}{*}{Val} & \textbf{context} & Şark meselesi temel hatlarıyla iki önemli süreçten oluşmaktadır. Bunlardan birincisi \colorbox{green}{'1071-1683' yılları arasındaki Şark Meselesi'dir}. Bu tarihler arasında Avrupa, Türklere karşı savunmaya geçmiş, Türkler ise fetihlere hız kazandırarak akıncılarını Avrupa topraklarına göndermiştir. Bu ilk süreç olan Şark Meselesinde temel amaç Türkleri Anadolu’ya sokmamak, Türklerin Anadolu’daki ilerleyişini durdurmak ve Türklerin Rumeli’ye girişini engellemektir. İstanbul’un Türkler tarafından fethini engellemek isteyen Avrupa devletleri, Türklerin Balkanlar üzerinden Avrupa içlerine doğru ilerleyişine engel olmak için birçok politikalar izlemişler ve bu politikalarda Şark Meselesinin ilk planlı durdurma safhasını oluşturmaktadır.\\
    & \textbf{generated question} & \textbf{Şark Meselesi'nin temel hatlarıyla iki önemli süreçten biri ne zamandır?}\\
    & \textbf{gold answer} & '1071-1683' yılları arasındaki Şark Meselesi'dir.\\
    \hline
    \rowcolor{Gray}
    \multirow{7}{*}{Val} & \textbf{context} & Maksimum Güç Noktası İzleyicisi (MPPT): Panellerden gelen enerjiyi düzenleyerek sürekli en yüksek güç üretimini sağlayan bir cihazdır. Bunu yapmak için akımı değiştirmektedir. Burada çıkan akım ile batarya şarj edilmektedir. Yarış sırasında en yüksek \colorbox{green}{15A} şarj değeri görülmüştür. Bu cihaz sayesinde akülerimizi daha kısa sürede şarj edecek akımı seçebiliyoruz.\\
    \rowcolor{Gray}
    & \textbf{generated question} & \textbf{Maximum Güç Noktası İzleyicisi akülerimizi daha kısa sürede şarj edecek hangi akımı değiştirmektedir?}\\
    \rowcolor{Gray}
    & \textbf{gold answer} & 15A\\
    \multirow{16}{*}{Val} & \textbf{context} & EK 33. Madde 1. paragraf: Ulusal egemenlik kapsamındaki uydu yörünge pozisyonlarının hakları, yönetimi ve işletme yetkisine sahip olmak ve bununla ilgili yükümlülükleri yerine getirmek, adına kayıtlı ve diğer operatörlere ait uyduları işletmeye vermek ya da verilmesini sağlamak, bu uyduları işletmek, ulusal ve yabancı operatörlere ait uydular üzerinden haberleşme ve iletim alt yapısını kurmak, işletmek ve ticarî faaliyette bulunmak üzere, bu Kanun ile kuruluş ve tescile ilişkin hükümleri hariç olmak üzere 6762 sayılı Türk Ticaret Kanunu ve özel hukuk hükümlerine tâbi, Türksat Uydu Haberleşme ve İşletme Anonim Şirketi (Türksat A.Ş.) unvanı altında bir anonim şirket kurulmuştur.EK 33. Madde 3. paragraf: Türksat A.Ş.'nin hisselerinin tamamı Hazine Müsteşarlığına aittir. Ancak, Hazine Müsteşarlığının mülkiyet hakkı ile kâr payı hakkına halel gelmemek ve kamunun pay sahipliğinden kaynaklanan bütün malî hakları Hazine Müsteşarlığında kalmak kaydıyla, Hazine Müsteşarlığının Türksat A.Ş.'deki pay sahipliğine dayanan oy, yönetim, temsil, denetim gibi hak ve yetkileri Ulaştırma Bakanlığı tarafından kullanılır. \colorbox{green}{800 civarında} televizyon ve radyo data yayını vardır. Şirket; televizyon ve radyo yayıncılığı, Kablo TV hizmeti, internet bağlantısı, veri transferi, VoIP gibi hizmetleri sunmaktadır.\\
    & \textbf{generated question} & \textbf{Türksat'ın kaç adet uydu yapısı vardır?}\\
    & \textbf{gold answer} & 800 civarında\\
    \hline
    \end{tabular}%
    \end{adjustbox}
    \label{tab:inconsistent-generation-samples}
\end{table}

\end{document}